\documentclass[letterpaper, 10 pt, conference]{ieeeconf}  

\IEEEoverridecommandlockouts                              

\usepackage{graphicx}
\usepackage{url}
\usepackage{graphics} 
\usepackage{amsmath} 
\usepackage{amssymb}  

\title{\LARGE \bf
Differentiable Boustrophedon Paths That Enable Optimization Via Gradient Descent}

\author{Thomas Manzini$^{*}$ and Robin Murphy$^{*}$
\thanks{$^{*}$Texas A\&M University, College Station, TX, USA. tmanzini@tamu.edu, robin.r.murphy@tamu.edu }
}

\begin{document}

\maketitle
\thispagestyle{empty}
\pagestyle{empty}

\begin{abstract}

This paper introduces a differentiable representation for the optimization of boustrophedon path plans in convex polygons, explores an additional parameter of these path plans that can be optimized, discusses the properties of this representation that can be leveraged during the optimization process and shows that the previously published attempt at optimization of these path plans was too coarse to be practically useful. 
Experiments were conducted to show that this differentiable representation can reproduce scores from traditional discrete representations of boustrophedon path plans with high fidelity. 
Finally, optimization via gradient descent was attempted but found to fail because the search space is far more non-convex than was previously considered in the literature.
The wide range of applications for boustrophedon path plans means that this work has the potential to improve path planning efficiency in numerous areas of robotics, including mapping and search tasks using uncrewed aerial systems, environmental sampling tasks using uncrewed marine vehicles, and agricultural tasks using ground vehicles, among numerous others applications.

\end{abstract}

\section{Introduction}
Boustrophedon path plans involve traversing an area of interest in straight lines with fixed spacing.
These straight lines are referred to as transects because they transect the area of interest.
These path plans are used extensively in aerial survey, environmental monitoring, and search tasks because they guarantee coverage over an area of interest.

\begin{figure}
  \centering
  \includegraphics[scale=0.45]{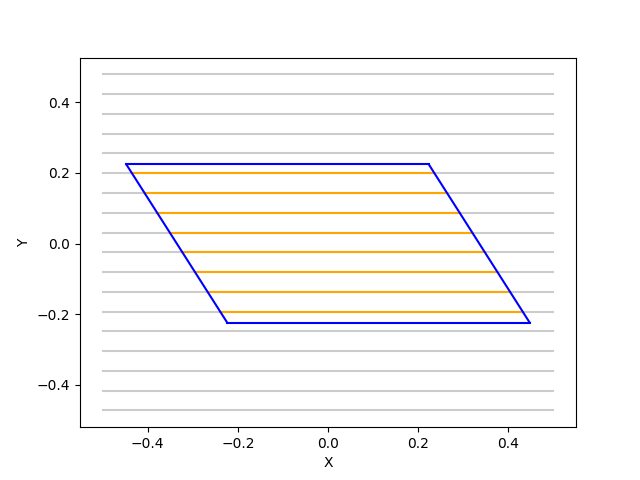}
  \vspace{0.075in}
  \centerline{(a)}
  \includegraphics[scale=0.45]{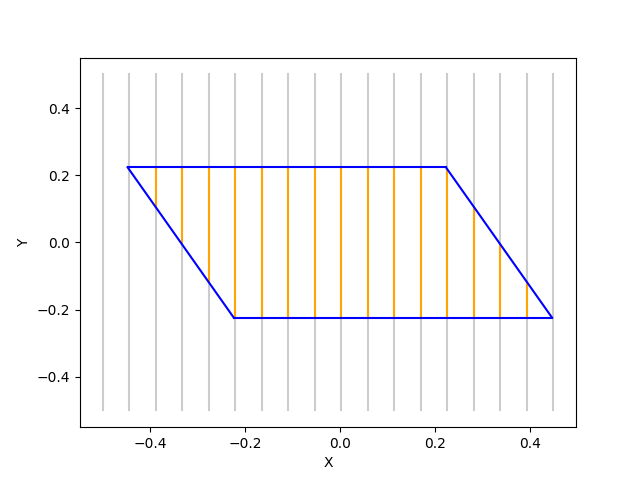}
  \vspace{0.075in}
  \centerline{(b)}
  \caption{Transects discretely generated within a parallelogram. Transects covering the polygon are shown in yellow. Some may consider the transects in (a) superior to (b) as the same space is covered using a smaller count of uniform-length transects.}
  \label{discrete_transects}
\end{figure}

Typically, executors of these path plans start on an edge-most transect that is followed until it terminates at the end of the area of interest, whereby the next proximal transect is selected and followed in the opposite direction.
This process continues until all transects have been followed.
An example of these transects is depicted in yellow in Figure \ref{discrete_transects} contained within an area of interest defined using a blue parallelogram.

The literature to date represents this problem in a discrete manner by computing transects that start and end at fixed positions along the perimeter of the polygon that defines the area of interest.
This approach is intuitive for robots that require discrete waypoints for use in path execution.

Recent advances in convex and non-convex optimization in machine learning have shown the effectiveness of optimization via gradient descent-based methods \cite{sejnowski2020unreasonable}; however, this discrete representation, while convenient computationally and operationally, precludes these techniques unless using encumbering gradient approximation methods. At the same time, introducing a differentiable system for optimizing boustrophedon path plans would enable the transmission of gradients into other differentiable systems, like neural networks, which are currently unable to optimize these path plans directly. 

This paper bridges this gap by presenting a numerically stable and mathematically differentiable representation for boustrophedon paths over convex polygons that enables the optimization of these paths via gradient descent\footnote{Source code is available at \url{https://github.com/TManzini/DifferentiableBoustrophedonPathPlans}}.
It will go on to show that this representation approximates the discrete representation with a high degree of fidelity, reveals that the optimization space is more complex than previously discussed, and requires search techniques that are far finer than the 10-degree grid search suggested in \cite{smith2022path}.
Finally, it will explore the properties of this differentiable representation that can mitigate, but not fully alleviate, the issues of non-convex optimization in this space.

In addition, this paper explores the optimization of an additional parameter of boustrophedon paths. The most relevant previous work only explored the optimization of the angle of the transects \cite{smith2022path}. However, this paper shows that shifting transects laterally also impacts the fitness of a particular path solution and that this parameter can be differentiated in the same manner as the transect angle.

\section{Related Work}

The general areas in which this work sits are well-developed. Boustrophedon path planning and its associated coverage guarantees have been extensively studied \cite{choset1998coverage, choset2000coverage, choset2001coverage} for decades. Numerous papers, too many to cite, have explored optimization via gradient descent. Recent advances in high dimensional optimization using gradient descent have fueled, in part, significant advances in machine learning \cite{ruder2016overview, sejnowski2020unreasonable}. What differs from previous work is this paper’s focus on optimizing the length and orientation of transects.

Optimization of boustrophedon path plans has been explored before. Boustrophedon path plans are generally optimized to guarantee coverage while minimizing the total path length. However, other optimizations have been considered.
For example, selecting which transect to execute next has been explored and reformulated as a variant of the traveling salesman problem \cite{bahnemann2021revisiting} and as a genetic search problem \cite{yuan2022global}. However, these approaches take the transects as a given and only seek to optimize the transitions between transects in non-holonomic vehicles. This is not an issue for holonomic vehicles as they have no turning rate constraints and can proceed directly to the next transect.

However, optimizing the locations and orientations of transects has been less studied. This is important because minimizing the path length may not be an appropriate optimization in all cases. 
The most relevant past work in this regard is \cite{smith2022path}, which explored optimizing the angle at which a set of transects were generated for convex and concave polygons.
The objective was to maximize the average length of the transects while minimizing the variance of those same transect lengths. The thought is that for sonar mapping tasks, marine vehicles would benefit from having long transects for platform stability, with minimal variance in length, to ease sensor data interpretation.
This work showed that certain orientations for these transects are favorable to others and suggested that these orientations can be optimized using grid search. They suggested using search intervals of 10 degrees to examine transect orientation.
The authors evaluated the effectiveness of this method over a series of polygons generated by uncrewed marine vehicle operators in the field.  
However, it should be noted that \cite{smith2022path} only explores the optimization of a single parameter---transect orientation---on boustrophedon path fitness.

Beyond boustrophedon path planning, coverage-based path planning has explored numerous other path planning strategies \cite{cabreira2019survey}, including optimizations for other axes such as time \cite{dong2023time} and failure tolerance in multi-robot systems \cite{ishat2021failure}.

\section{Problem Formulation}

The problem formulation is presented first as the traditional discrete approach for generating these path plans to establish a base case, followed by the differentiable approach proposed here.

\subsection{Discrete Approaches}

As has been discussed, boustrophedon path plans are constructed out of a set of transects arranged at fixed intervals along the perimeter of the polygon.
In the discrete setting, these transects are generated by iterating over the perimeter of the polygon and generating transects as appropriate.
These transects are then connected along the edge of the polygon in an alternating fashion.

An example of such a discrete set of transects is shown in Figure \ref{discrete_transects}. 
In this figure, a field of candidate transects is placed in the same space as the polygonal area of interest.
The final transects are then determined by computing the intersection of these candidate transects with the polygon.
Thus, transects are precisely the dimension of the polygon and have discrete endpoints. 

Past work \cite{smith2022path} scores transect orientations based on the count and dimension of the transects. Specifically, this work proposes maximizing equation \ref{discrete_opt_eq}.

\begin{equation}
    f(\theta) = a \cdot (m'_\theta ) + b  \cdot (1-COV_\theta)
    \label{discrete_opt_eq}
\end{equation}

Where $ m’_\theta$ is the average transect length, $COV_\theta$ is the standard deviation of transect lengths, and $a$ and $b$ are hyperparameters that weight the optimization over angles $\theta$. 

This formulation precludes differentiation because the count of transects at each angle $\theta$, used to compute the mean and standard deviation, is a discontinuous function that cannot be differentiated. 

\subsection{Differentiable Approach}
\label{diff_approach}

\begin{figure*}
  \includegraphics[width=\textwidth]{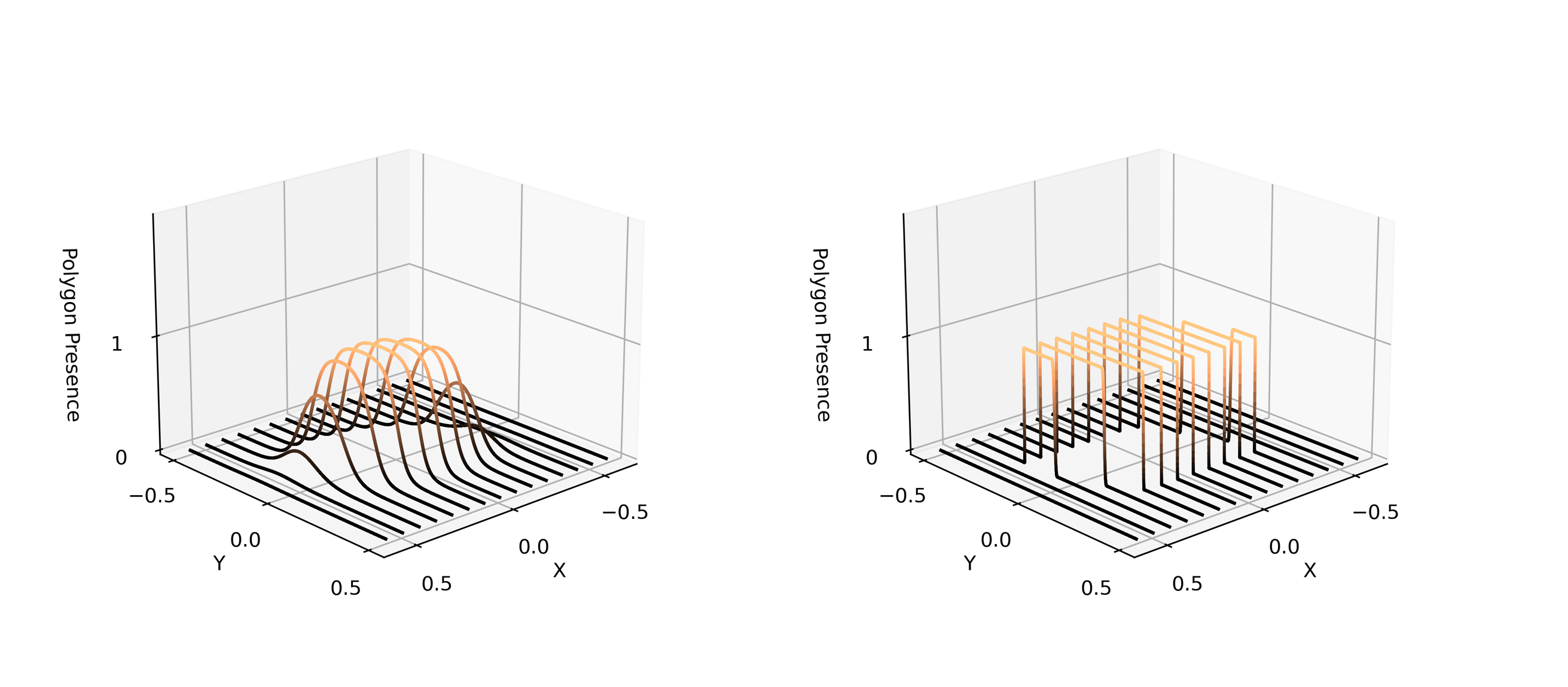}
  \caption{Transect lines passing through the parallelogram from Figure \ref{discrete_transects} using the differentiable formulation described. The z-axis corresponds to an indicator function stating if a particular point is contained within the parallelogram. Transect lengths are approximated based on the integral of these lines. Left shows Temperature=17, right shows Temperature=100.}
  \label{transect_lines_by_temp}
\end{figure*}

The proposed differentiable approach also requires a definition for a polygon and transects but differs in its implementation from the discrete approach. This section defines how to differentiably evaluate a given transect parameterization. 

\subsubsection{Polygon Definition}
\label{polygon_def}

To come up with a differentiable function for transect fitness with respect to a given polygon, a definition of a polygon that supports differentiation must be utilized.
To do this, a formulation is developed leaning on the universal approximation theorem from artificial neural networks \cite{cybenko1989approximation}. This formulation utilizes the sum of the sigmoid of several linear functions to approximate the bounds of a polygon. More specifically, each face of a polygon is approximated using a linear function $W^TX + B$ where the matrix $W$ describes the slope of the edge of a polygon, $B$ describes an offset to align the edge with the rest of the polygon, and $X$ describes the point to be evaluated. The procedure for constructing $W$ and $B$ from a set of two points is straightforward. First, the polygon is resized such that its maximal dimension equals one and is centered on the origin. Then, the $W$ and $B$ parameters of the polygon edges are defined following equation \ref{point2face}. 

\begin{equation}
\begin{aligned}
    Given \; (x_1, y_1), (x_2, y_2) \\\\
    X_e = \begin{bmatrix}
    x_1 & 1 \\
    x_2 & 1
    \end{bmatrix}, 
    Y_e = \begin{bmatrix}
    y_1 & 1 \\
    y_2 & 1
    \end{bmatrix}\\
    C_e = \begin{bmatrix}
    x_1 & y_1 \\
    x_2 & y_2
    \end{bmatrix}\\\\
    W = [Det(X_e), Det(Y_e)]\\
    B = Det(C_e)\\
\end{aligned}
\label{point2face}
\end{equation}

Now, each face of the polygon defined in this way is passed to a sigmoid function ($\sigma$) to compress it between 0 and 1, as shown in equation \ref{face2sigmoid}. A hyperparameter Temp, referred to as Temperature, controls the steepness of this sigmoid function. Temperature is defined as having a value (0, $+\infty$). Additional details of this hyperparameter are discussed in section \ref{discuss_temp}. 

\begin{equation}
\begin{aligned}
v = W^TX + B\\
\sigma(v, Temp) = \frac{1}{1+e^{-v \cdot Temp}}
\end{aligned}
\label{face2sigmoid}
\end{equation}

With this function, it can be determined if a given point is on one side of a given polygon’s face. A product is taken over all the polygon edges $F$ in equation \ref{sigmoid2polygon} to determine if a given point is contained within all faces.

\begin{equation}
\begin{aligned}
P(X, F, Temp) = \prod_{(W, B) \in F} \sigma(W^TX+B, Temp)
\end{aligned}
\label{sigmoid2polygon}
\end{equation}

This behaves as an approximate indicator function of whether the point $X$ is contained within the given polygon faces $F$; however, it is not numerically stable because high fidelity approximations require exceedingly high temperatures (Temp $\ge10^5$). To combat this, the value of equation \ref{sigmoid2polygon} is passed to another sigmoid function as shown in equation \ref{sigmoid2polygon_compress}. This increases approximation fidelity at low temperatures, thereby preserving numerical stability.

\begin{equation}
\begin{aligned}
P_\sigma(X, F, Temp) = \sigma(P(X, F, Temp) - 0.5, Temp)
\end{aligned}
\label{sigmoid2polygon_compress}
\end{equation}

Equation \ref{sigmoid2polygon_compress} represents the final function used to approximate convex polygons. This function is entirely differentiable and provides high fidelity while being numerically stable. It should also be noted that this polygon formulation limits this approach to convex polygons, as it requires a point to be contained by all faces of the polygon, which will only be true when there is convexity. However, \cite{cybenko1989approximation} shows that extending this approach to concave polygons is possible.

\subsubsection{Transect Definition}

A set of lines $\gamma$ at equal width is defined to approximate the transects. As the polygon was rescaled such that the maximum dimension was between -0.5 and 0.5, these lines run parallel to the Y axis from -0.5 to 0.5 and are repeated along the X axis at the predefined, but rescaled, spacing covering -0.5 to 0.5. Example transects are shown in figure \ref{transect_lines_by_temp} and are evaluated by the polygon representation defined in section \ref{polygon_def} at different temperatures.

The critical item that must be calculated in this setting is the transect length. 
The integral of the transect passing through the polygon can be computed to approximate the transect length. This is shown in equation \ref{transect_len}. Defining the transects to be parallel to an axis simplifies this integral as it allows $x$ to remain constant, meaning only $y$ needs to be integrated.

\begin{equation}
\begin{aligned}
\int_{-0.5}^{0.5} P_\sigma([x, y], F, Temp) dy
\end{aligned}
\label{transect_len}
\end{equation}

This allows us to compute the length of a single transect. Downstream differentiable scoring functions may then consume this information for optimization purposes.

At the time of writing, a closed-form solution to the integral described in equation \ref{transect_len} is not known. Thus, this integral is approximated using the trapezoidal rule.

\subsubsection{Scoring Function Definition}

With a differentiable definition of a polygon and a formulation of transects that allows for the approximation of a transect length given a polygon, the optimization of those transects may now be discussed.
With a differentiable approximation of the length of a transect, a wide variety of scoring functions can be imagined. In keeping with the literature, the fitness function from \cite{smith2022path} in equation \ref{discrete_opt_eq} is adapted to this setting.

To compute the average length of all transects, it must be determined if a given transect in $\gamma$ has length. To do this, an additional sigmoid function is used as an indicator function for transects that have an approximated length \>0. This is done as described in equation \ref{transect_shown}.

\begin{equation}
\begin{aligned}
S(x, F, Temp) = \frac{\sigma( \int_{-0.5}^{0.5} P_\sigma([x, y], F, Temp) dy)- 0.5}{0.5}
\end{aligned}
\label{transect_shown}
\end{equation}

The constants $0.5$ in equation \ref{transect_shown} are to scale the indicator function. This function will now equal 0 when the transect at $x$ has no length. From equation \ref{transect_shown}, the mean and standard deviations of transect with lengths \>0 can be approximated using the approximated transect lengths and an indicator function detailing whether they were shown.

It is worth noting that while this work focuses on a differentiable implementation of the fitness function presented in \cite{smith2022path}, the problem formulation described in \ref{diff_approach} is not limited to this fitness function. 
This approach enables a variety of fitness functions that may be useful depending on the needs at the time. For example, in keeping with much of the literature, it may be worthwhile to minimize the total length of all transects; this is possible under this formulation by simply summing the approximated length of all transects. 
In fact, any function that has a defined derivative may be utilized. Readers are encouraged to adapt the fitness function to fit their needs.

\subsubsection{Computation of Gradients}

As discussed in the introduction, there are two parameters for which gradients are computed: angle and x-offset.

In the case of angle, a simplification is made compared to \cite{smith2022path}. Instead of rotating the transects, the polygon is rotated. This simplifies the integral function defined in equation \ref{transect_len}. This is done through a rotational transform applied to the $W$ and $B$ terms to rotate them about the origin. Gradients are then computed on the angle passed to this transform.

In the case of x-offset, this is done by adding a parameter to the $x$ term in equation \ref{transect_len} and computing the gradient on that additional term.

More practically, any popular auto-differentiation package can readily compute the gradients necessary for optimization. PyTorch \cite{NEURIPS2019_9015} was utilized in the case of this work.

\begin{figure}
  \includegraphics[width=\columnwidth]{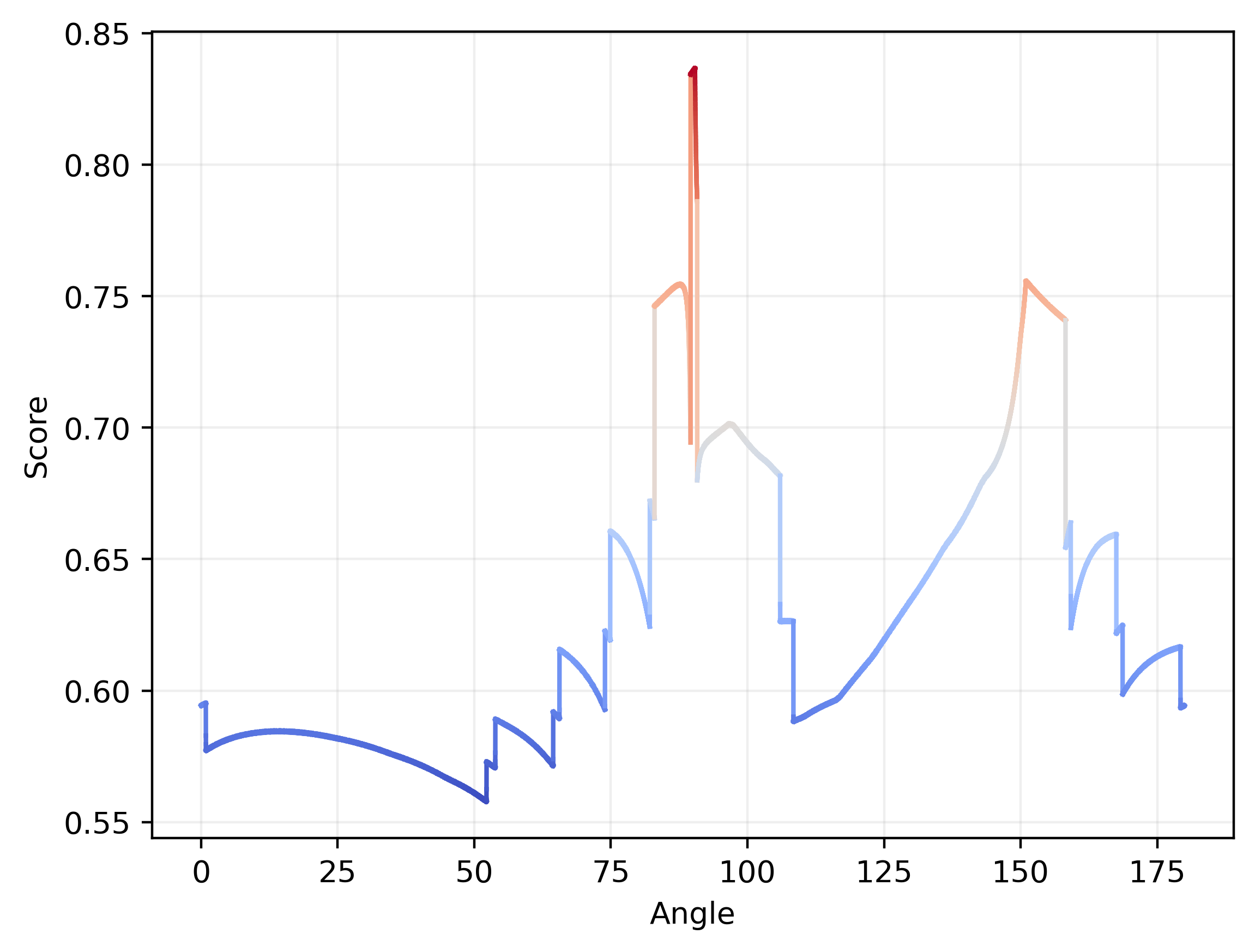}
  \caption{The score surface of the differentiable formulation based on the parallelogram and transects, shown in Figure \ref{discrete_transects}. Here, the angle varies, and the x-offset is fixed at 0.0 and Temperature at 10000.}
  \label{surface_slice}
\end{figure}

\section{Comparison with Discrete Representations}
\label{discrete_comparison}
An obvious question that follows this complex mathematical machinery is whether it sufficiently approximates the discrete formulation. 
To answer this question, an experiment was conducted where random orientations, x-offsets, transect spacings, and convex polygons with sides $[3,10)$ were generated.
These parameters were then passed to a discrete system where the transect lengths and subsequent fitness scores were computed precisely and the differentiable system where the fitness scores were approximated.
Both of these fitness scores were measured, and absolute differences were computed.
In this context, the discrete implementation is considered truth, and any deviation from the discrete score is considered error.
The fitness function from \cite{smith2022path} and shown in equation \ref{discrete_opt_eq} was used, resulting in a bounded score between 0 and 1. 
For additional context, the Temperature and the number of points used to approximate the integral in equation \ref{transect_len} were varied, and 1000 samples were taken for each variation.
The results of this experiment are shown in Table \ref{tab:error}.

\begin{table}[]
\centering
\begin{tabular}{|l|ccc|}
\hline
Error   & \multicolumn{1}{l|}{100 PPT} & \multicolumn{1}{l|}{1000 PPT} & \multicolumn{1}{l|}{10000 PPT} \\ \hline
Temperature=1     & 0.50798                      & 0.50795                       & 0.50408                        \\ \cline{1-1}
Temperature=10    & 0.08480                      & 0.08369                       & 0.08367                        \\ \cline{1-1}
Temperature=100   & 0.03090                      & 0.03001                       & 0.03167                        \\ \cline{1-1}
Temperature=1000  & 0.00374                      & 0.00271                       & 0.00306                        \\ \cline{1-1}
Temperature=10000 & 0.00203                      & 0.00055                       & 0.00016                        \\ \hline
\end{tabular}
\caption{Average error between discrete and differentiable scores when comparing Temperature and points per transect (PPT).}
\label{tab:error}
\end{table}

Table \ref{tab:error} shows that the error decreases as the Temperature and number of points used to approximate the integral grow. The minimum error observed was 0.00016 on a fitness function bounded between 0-1. It is argued that this corresponds to a high degree of fidelity and is sufficiently accurate for the optimization tasks presented in \cite{smith2022path}. 

In addition, it should be noted that while increasing the Temperature beyond the ranges of this table may increase precision, it may also lead to numerical instability due to overflow of the floating point representation. At the same time, while increasing the points per transect does increase the fidelity of the representation by increasing the resolution of equation \ref{transect_len}, it also increases run-time.

\section{Discussion of the Optimization Space}

With reasonable parity between the discrete and differentiable representations, the discussion turns to the nature of the optimization space.
Taking the example shown in figure \ref{discrete_transects} and a differentiable implementation of the target function in Equation \ref{discrete_opt_eq}, figure \ref{surface_slice} shows the optimization surface generated when varying the angle and keeping the x-offset fixed. Even in this simple polygon with few transects, there is intense non-convexity. This non-convexity is not a function of the differentiable representation, as there is very close parity between the discrete and differentiable representations at this scale. Instead, this non-convexity is because fewer transects are visible in some orientations of the polygon than in others. As the polygon rotates, some transects leave the volume of the polygon. This leads to decreases in fitness as the transects become smaller and then a rapid increase in fitness after the small outlier transect leaves the polygon.

\begin{figure}
  \includegraphics[width=\columnwidth]{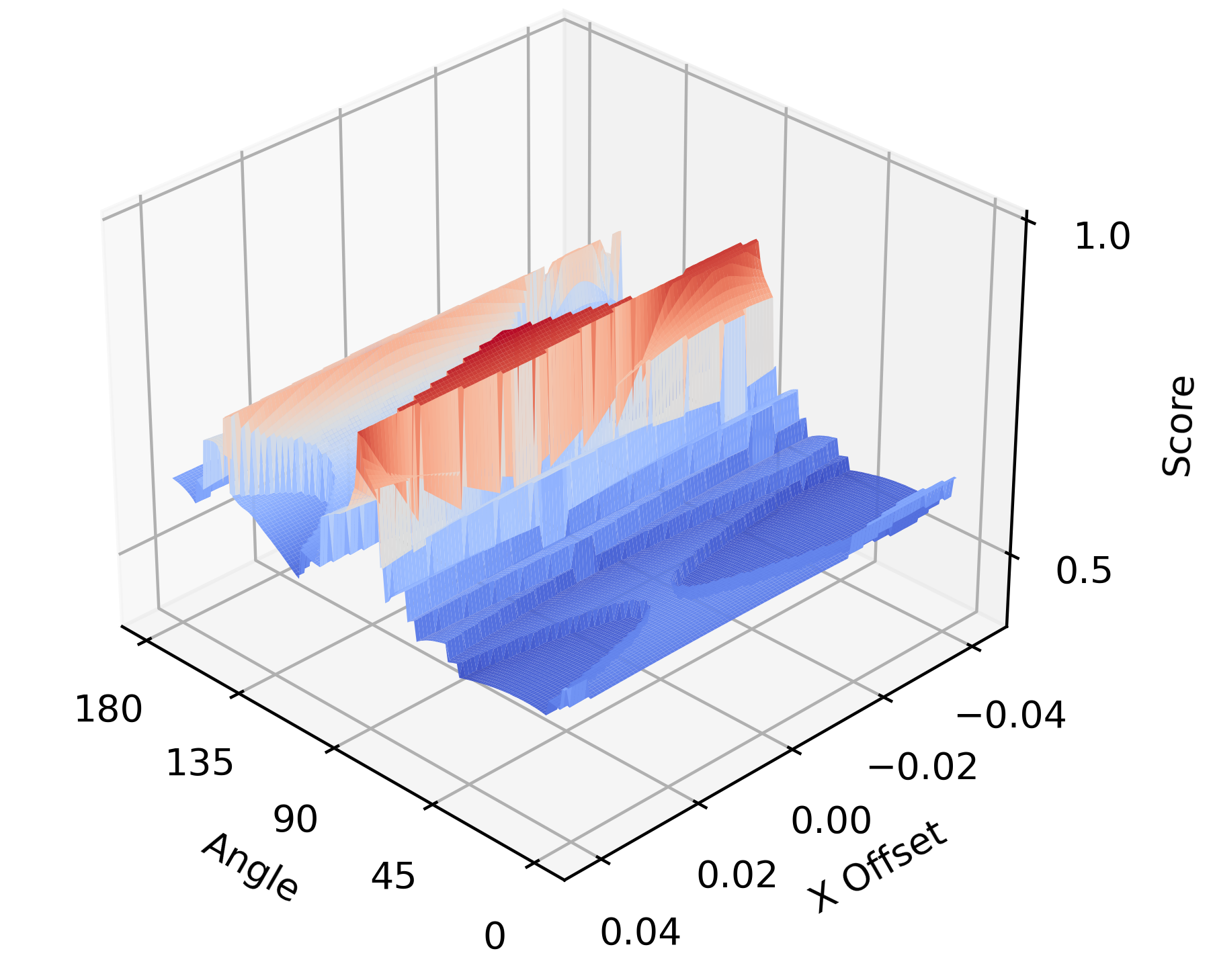}
  \caption{The 2D score surface is described by the differentiable formulation based on the parallelogram and transects, shown in Figure \ref{discrete_transects}. Here, the angle and the x-offset are varied. The slice where x-offset is equal to 0.0 is displayed in Figure \ref{surface_slice}.}
  \label{surface_full}
\end{figure}

Further optimization surface complexity is observed when varying the x-offset variable. In figure \ref{surface_full}, again utilizing the polygon and settings from figure \ref{discrete_transects}, the optimization surface of both the angle and the x-offset is shown. While the complexity of the x-offset axis appears lesser than that of the angle axis, it no doubt impacts the shape of the optimization surface, meaning that it needs to be considered in future optimization efforts.

The intense non-convexity of this optimization surface means that future efforts in searching these spaces need to operate at a far more fine-grained resolution than the 10-degree interval suggested in \cite{smith2022path}. 

Two main factors drive this non-convexity: transect spacing and polygon aspect.
In the case of transect spacing, as transect spacing decreases and the transect count increases, the non-convexity of the function increases because the rate at which transects will be entering and exiting the polygons volume will increase. 
In the case of polygon aspect, polygons with a high aspect ratio will have more non-convex loss surfaces because as the polygon rotates, more transects will enter and exit the polygon’s volume. While a circle, or a shape with an aspect ratio of 1, would have a uniform loss surface as any orientation would be equally as fit.
These two factors play together to create the non-convex nature of the optimization surface.

\section{Discussion of Temperature}
\label{discuss_temp}

Temperature is the only hyperparameter that is introduced in this representation, and as a result, its properties are worth discussing.
The Temperature can be seen as a parameter that governs the discreteness of transect lengths. 
At high temperatures, the transect lengths are approximated with greater fidelity and, therefore, discreteness than at lower temperatures.
When viewing a single transect, increasing the Temperature can be seen as discretizing the transect; however, the effect is more intuitive when considering the whole polygon.
From the perspective of the polygon, lowering the temperature “melts” the polygon, and raising the Temperature solidifies it; as shown in figure \ref{transect_lines_by_temp}.

\begin{figure}
  \includegraphics[width=\columnwidth]{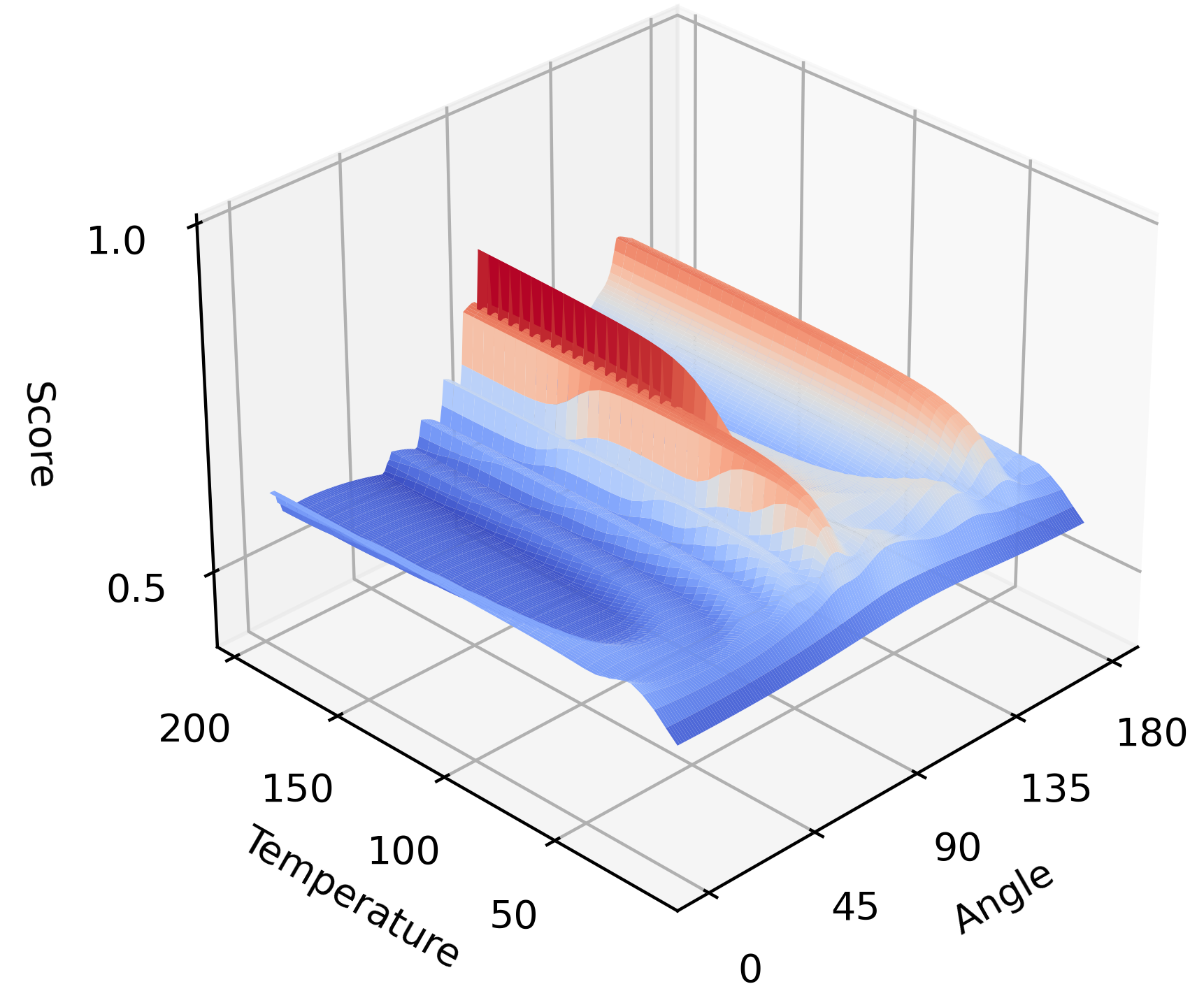}
  \caption{The 2D score surface as described by the differentiable formulation based on the parallelogram and transects, shown in Figure \ref{discrete_transects}. Here, the angle and Temperature are varied. While this figure only shows temperatures less than 200, the slice where the Temperature is equal to 10000 is displayed in Figure \ref{surface_slice}.}
  \label{surface_temp}
\end{figure}

From the perspective of the loss surface, it has the same effect. The loss surface becomes far smoother and undulating at lower temperatures than high temperatures. This is shown in figure \ref{surface_temp}, where the Temperature is varied in conjunction with the angle. It should be noted that while gradients can also be computed for the Temperature, it is not recommended to optimize this parameter directly as it impacts the fidelity of the representation, and optimizers may find themselves trapped in temperature-based optima.

However, it may be beneficial to vary the Temperature depending on the progress of the optimization. 
For example, one could perform gradient descent in a low-temperature setting initially, leveraging the smoother loss surface and then increasing the Temperature and fidelity, in turn, to refine the optima later in the optimization process.

\section{Optimization via Gradient Descent}

As discussed, the loss surface is intensely non-convex, with numerous local optima and only one global optima. This presents a challenging optimization space, as it can be impossible to know if you have reached the global minima. 

To highlight this difficulty, an experiment was conducted that compared grid search and a gradient descent approach when optimizing 1000 randomly generated convex polygons, generated in the manner as the experiment in section \ref{discrete_comparison}. In this experiment, gradient descent with a momentum of 0.8 was used\footnote{This formulation is not limited to traditional gradient descent and can support the gambit of optimizers discussed in the literature \cite{ruder2016overview}.}. It was found that grid search arrived at an optimum that was 0.13633 higher, on average, than the gradient descent-based approach. However, gradient descent occasionally converged to an optima greater than grid search. This is unsurprising given the non-convexity observed in this loss surface in figures \ref{surface_slice}, \ref{surface_full}, and \ref{surface_temp}.

This result highlights how much more non-convex this optimization space is than was previously thought. Further, it represents the well-known fact that optimization in non-convex spaces remains challenging, and the tradeoffs between gradient descent and grid search proceed in the usual way. This work is not a solution to the non-convexity presented by this problem but another tool that may be used to improve optimization efforts in this setting.

It is theorized that the best chance of arriving at the global optima efficiently in this setting will be through a combination of grid search and gradient descent, where grid search will be utilized to arrive at a parameterization near the global optima, which gradient descent can further refine.

\section{Conclusions}

While optimization via gradient descent remains challenging due to the intense non-convexity found in the loss surface, the analysis in this paper demonstrates that optimization via gradient descent can provide utility as a mechanism to refine further solutions generated via grid search.
The results show that a differentiable implementation of boustrophedon path plans over convex polygons approximates the discrete formulation commonly used with a high degree of fidelity that can be tuned depending on the needs of the optimization problem.
It also shows that the optimization surface is far more non-convex than previously discussed in the literature, motivating the need for far more fine-grained search methods than previously discussed. 
From a broader perspective, this work contributes to understanding the details of the optimization space, mainly by providing an explanation for the non-convexity, the utility of the temperature hyperparameter in the differentiable representation, and the use of gradient descent to optimize path plans.

Future work will focus on three elements: first, developing a closed-form solution to the integral in equation \ref{transect_len}; second, refining optimization strategies that utilize gradient descent and grid search to arrive at more precise optima; third, exploring extensions of this representation that capture concave areas of interest. Developing a closed-form solution to the integral in equation \ref{transect_len} would result in a speed-up equivalent to the number of points per transect used to calculate the integral empirically while also increasing fidelity. 

\addtolength{\textheight}{-12cm}

\section*{Acknowledgements}
Acknowledgments are given to Trey Smith, who helped validate some of the initial findings, and Hydronalix, who provided funding, in part, for this work.

\bibliographystyle{IEEEtran}
\bibliography{root}

\begin{thebibliography}{10}
\providecommand{\url}[1]{#1}
\csname url@samestyle\endcsname
\providecommand{\newblock}{\relax}
\providecommand{\bibinfo}[2]{#2}
\providecommand{\BIBentrySTDinterwordspacing}{\spaceskip=0pt\relax}
\providecommand{\BIBentryALTinterwordstretchfactor}{4}
\providecommand{\BIBentryALTinterwordspacing}{\spaceskip=\fontdimen2\font plus
\BIBentryALTinterwordstretchfactor\fontdimen3\font minus
  \fontdimen4\font\relax}
\providecommand{\BIBforeignlanguage}[2]{{%
\expandafter\ifx\csname l@#1\endcsname\relax
\typeout{** WARNING: IEEEtran.bst: No hyphenation pattern has been}%
\typeout{** loaded for the language `#1'. Using the pattern for}%
\typeout{** the default language instead.}%
\else
\language=\csname l@#1\endcsname
\fi
#2}}
\providecommand{\BIBdecl}{\relax}
\BIBdecl

\bibitem{sejnowski2020unreasonable}
T.~J. Sejnowski, ``The unreasonable effectiveness of deep learning in
  artificial intelligence,'' \emph{Proceedings of the National Academy of
  Sciences}, vol. 117, no.~48, pp. 30\,033--30\,038, 2020.

\bibitem{smith2022path}
T.~Smith, S.~Mukhopadhyay, R.~R. Murphy, T.~Manzini, and I.~Rodriguez, ``Path
  coverage optimization for usv with side scan sonar for victim recovery,'' in
  \emph{2022 IEEE International Symposium on Safety, Security, and Rescue
  Robotics (SSRR)}.\hskip 1em plus 0.5em minus 0.4em\relax IEEE, 2022, pp.
  160--165.

\bibitem{choset1998coverage}
H.~Choset and P.~Pignon, ``Coverage path planning: The boustrophedon cellular
  decomposition,'' in \emph{Field and service robotics}.\hskip 1em plus 0.5em
  minus 0.4em\relax Springer, 1998, pp. 203--209.

\bibitem{choset2000coverage}
H.~Choset, ``Coverage of known spaces: The boustrophedon cellular
  decomposition,'' \emph{Autonomous Robots}, vol.~9, pp. 247--253, 2000.

\bibitem{choset2001coverage}
------, ``Coverage for robotics--a survey of recent results,'' \emph{Annals of
  mathematics and artificial intelligence}, vol.~31, pp. 113--126, 2001.

\bibitem{ruder2016overview}
S.~Ruder, ``An overview of gradient descent optimization algorithms,''
  \emph{arXiv preprint arXiv:1609.04747}, 2016.

\bibitem{bahnemann2021revisiting}
R.~B{\"a}hnemann, N.~Lawrance, J.~J. Chung, M.~Pantic, R.~Siegwart, and
  J.~Nieto, ``Revisiting boustrophedon coverage path planning as a generalized
  traveling salesman problem,'' in \emph{Field and Service Robotics: Results of
  the 12th International Conference}.\hskip 1em plus 0.5em minus 0.4em\relax
  Springer, 2021, pp. 277--290.

\bibitem{yuan2022global}
J.~Yuan, Z.~Liu, Y.~Lian, L.~Chen, Q.~An, L.~Wang, and B.~Ma, ``Global
  optimization of uav area coverage path planning based on good point set and
  genetic algorithm,'' \emph{Aerospace}, vol.~9, no.~2, p.~86, 2022.

\bibitem{cabreira2019survey}
T.~M. Cabreira, L.~B. Brisolara, and F.~J. Paulo~R, ``Survey on coverage path
  planning with unmanned aerial vehicles,'' \emph{Drones}, vol.~3, no.~1, p.~4,
  2019.

\bibitem{dong2023time}
D.~Dong, H.~Berger, and I.~Abraham, ``Time optimal ergodic search,''
  \emph{arXiv preprint arXiv:2305.11643}, 2023.

\bibitem{ishat2021failure}
M.~Ishat-E-Rabban and P.~Tokekar, ``Failure-resilient coverage maximization
  with multiple robots,'' \emph{IEEE Robotics and Automation Letters}, vol.~6,
  no.~2, pp. 3894--3901, 2021.

\bibitem{cybenko1989approximation}
G.~Cybenko, ``Approximation by superpositions of a sigmoidal function,''
  \emph{Mathematics of control, signals and systems}, vol.~2, no.~4, pp.
  303--314, 1989.

\bibitem{NEURIPS2019_9015}
A.~Paszke, S.~Gross, F.~Massa, A.~Lerer, J.~Bradbury, G.~Chanan, T.~Killeen,
  Z.~Lin, N.~Gimelshein, L.~Antiga, A.~Desmaison, A.~Kopf, E.~Yang, Z.~DeVito,
  M.~Raison, A.~Tejani, S.~Chilamkurthy, B.~Steiner, L.~Fang, J.~Bai, and
  S.~Chintala, ``Pytorch: An imperative style, high-performance deep learning
  library,'' in \emph{Advances in Neural Information Processing Systems
  32}.\hskip 1em plus 0.5em minus 0.4em\relax Curran Associates, Inc., 2019,
  pp. 8024--8035.

\end{thebibliography}

\end{document}